\documentclass{article}
\usepackage{ijcai25}
\usepackage{multirow} 
\usepackage{times}
\usepackage{soul}
\usepackage{url}
\usepackage[hidelinks]{hyperref}
\usepackage[utf8]{inputenc}
\usepackage[small]{caption}
\usepackage{graphicx}
\usepackage{amsmath}
\usepackage{amsthm}
\usepackage{booktabs}
\usepackage{algorithm}
\usepackage{amssymb}
\usepackage{algorithmic}
\usepackage{subcaption}
\usepackage[switch]{lineno}

\title{DANCE: Resource-Efficient Neural Architecture Search with \\Data-Aware and Continuous Adaptation}

\author{
Maolin Wang$^{1}$
\and
Tianshuo Wei$^{1}$\and
Sheng Zhang$^1$\and
Ruocheng Guo$^{2,}$\footnote{Corresponding authors.}\and\\
Wanyu Wang$^{1,*}$ \and
Shanshan Ye$^{3,*}$\and
Lixin Zou$^4$\and
Xuetao Wei$^5$\and
Xiangyu Zhao$^1$\\
\affiliations
$^1$City University of Hong Kong\\
$^2$Independent Researcher\\
$^3$Australian Artificial Intelligence Institute, University of Technology Sydney\\
$^4$Wuhan University\\
$^5$Southern University of Science and Technology\\
\emails
rguo.asu@gmail.com,
wanyuwang4-c@cityu.edu.hk,
shanshan.ye@student.uts.edu.au
}
\begin{document}

\maketitle

\begin{abstract}
Neural Architecture Search (NAS) has emerged as a powerful approach for automating neural network design. However, existing NAS methods face critical limitations in real-world deployments: architectures lack adaptability across scenarios, each deployment context requires costly separate searches, and performance consistency across diverse platforms remains challenging. We propose DANCE (Dynamic Architectures with Neural Continuous Evolution), which reformulates architecture search as a continuous evolution problem through learning distributions over architectural components. DANCE introduces three key innovations: a continuous architecture distribution enabling smooth adaptation, a unified architecture space with learned selection gates for efficient sampling, and a multi-stage training strategy for effective deployment optimization.
Extensive experiments across five datasets demonstrate DANCE's effectiveness. Our method consistently outperforms state-of-the-art NAS approaches in terms of accuracy while significantly reducing search costs. Under varying computational constraints, DANCE maintains robust performance while smoothly adapting architectures to different hardware requirements. The code and appendix can be found at https://github.com/Applied-Machine-Learning-Lab/DANCE.
\end{abstract}

\section{Introduction}

Neural Architecture Search (NAS) has revolutionized deep neural network design by automating the architecture optimization process~\cite{chen2022automated}. Despite its recent success, existing NAS methods face significant challenges in effectively modeling three key aspects: architecture adaptability~\cite{zhao2022adaptive}, deployment efficiency~\cite{zhu2023autogen}, and performance consistency across diverse scenarios~\cite{ren2021comprehensive,zhang2023autostl}. While current approaches can address certain aspects, they often fail to comprehensively tackle all these challenges simultaneously~\cite{cai2019once}.

In real-world deployment scenarios, neural architectures naturally exhibit three key characteristics: (1)~\textbf{Compute Constraints}: architectures need to adapt to varying computational constraints (e.g., an architecture suitable for GPU servers may be impractical on mobile devices)~\cite{howard2017mobilenets,liu2020automated,zhao2021autodim,DNS-Rec}, (2)~\textbf{Varied Requirements}: deployment scenarios vary significantly in their requirements (e.g., real-time applications demand low latency while offline tasks prioritize accuracy)~\cite{wang2019haq,zhu2023autogen}, and (3)~\textbf{Performance Variance}: model performance patterns often differ across deployment contexts (e.g., models optimized for one domain may underperform in others)~\cite{tan2019efficientnet,gao2023autotransfer}. These observations highlight the critical need for dynamic and adaptable neural architectures that can efficiently handle diverse deployment scenarios while maintaining robust performance~\cite{zhao2022adaptive}.

Based on these observations, neural architecture search systems need to address three fundamental challenges that require innovative solutions:
First, architecture flexibility needs to be effectively modeled. Current discrete search methods lack the ability to smoothly adapt architectures as deployment requirements change, necessitating expensive re-search for each scenario~\cite{zoph2016neural,zhao2022adaptive,zhu2023autogen,lin2022adafs}.
Second, deployment scenarios exhibit diverse computational constraints. While recent methods have shown promise in handling specific deployment contexts~\cite{wu2019fbnet,liu2020automated}, they struggle to maintain consistent performance across varying scenarios due to their rigid architecture designs~\cite{gao2023autotransfer,li2023automlp,song2022autoassign}.
Third, architecture search naturally involves multiple competing objectives. Despite their advantages in finding optimal architectures, existing methods typically process these objectives uniformly, failing to capture the complex trade-offs between performance metrics~\cite{elsken2019neural,zhao2021autoloss,zhang2023autostl,jin2021automated}.
These limitations of current approaches are particularly evident in practice: Pure evolutionary methods demonstrate sub-optimal efficiency in large-scale deployments~\cite{real2019regularized,liu2024autoassign+}, gradient-based approaches face significant scalability challenges~\cite{el2017scalable}, and while one-shot methods offer improved search efficiency~\cite{dong2023review,zhaok2021autoemb}, they not only show limited transfer capability but also lack mechanisms to effectively capture diverse deployment requirements~\cite{bender2018understanding,chen2022automated,li2023automlp}. This creates a clear need for a unified solution that can simultaneously address architecture adaptability and deployment efficiency while maintaining consistent performance.

To address these challenges, we propose DANCE (\textbf{D}ynamic \textbf{A}rchitectures with \textbf{N}eural \textbf{C}ontinuous \textbf{E}volution), a novel framework that synergistically combines continuous architecture evolution and distribution learning through a unified optimization approach. Unlike previous methods that focus on block-level components, DANCE operates directly on feature dimensions, providing a unified perspective that bridges the gap between NAS and pruning. The continuous evolution mechanism enables smooth architecture adaptation across different deployment scenarios, while the distribution learning component efficiently captures and models deployment patterns. By integrating data-aware dynamic selection and importance-based sampling at the feature level, this unified framework effectively leverages both architectural flexibility and deployment efficiency while maintaining computational tractability through selective sampling and adaptive optimization strategies.

Our main contributions are summarized as follows:
\begin{itemize}

\item We propose DANCE, a novel neural architecture search framework that learns continuous distributions over architectural components. This enables efficient architecture adaptation under varying computational budgets while maintaining high performance.

\item We design a unified architecture space with dynamic select gates that integrate batch-level features and layer-wise importance metrics. This mechanism enables flexible component selection and smooth architecture adaptation across different hardware constraints and deployment contexts.

\item Extensive experiments on five image datasets demonstrate that DANCE achieves consistent accuracy improvements over state-of-the-art methods while significantly reducing search costs. The results validate its effectiveness for practical applications with diverse computational requirements and resource constraints.

\end{itemize}

\section{Methodology}

\begin{figure}[t]
\centering
\includegraphics[width=\linewidth]{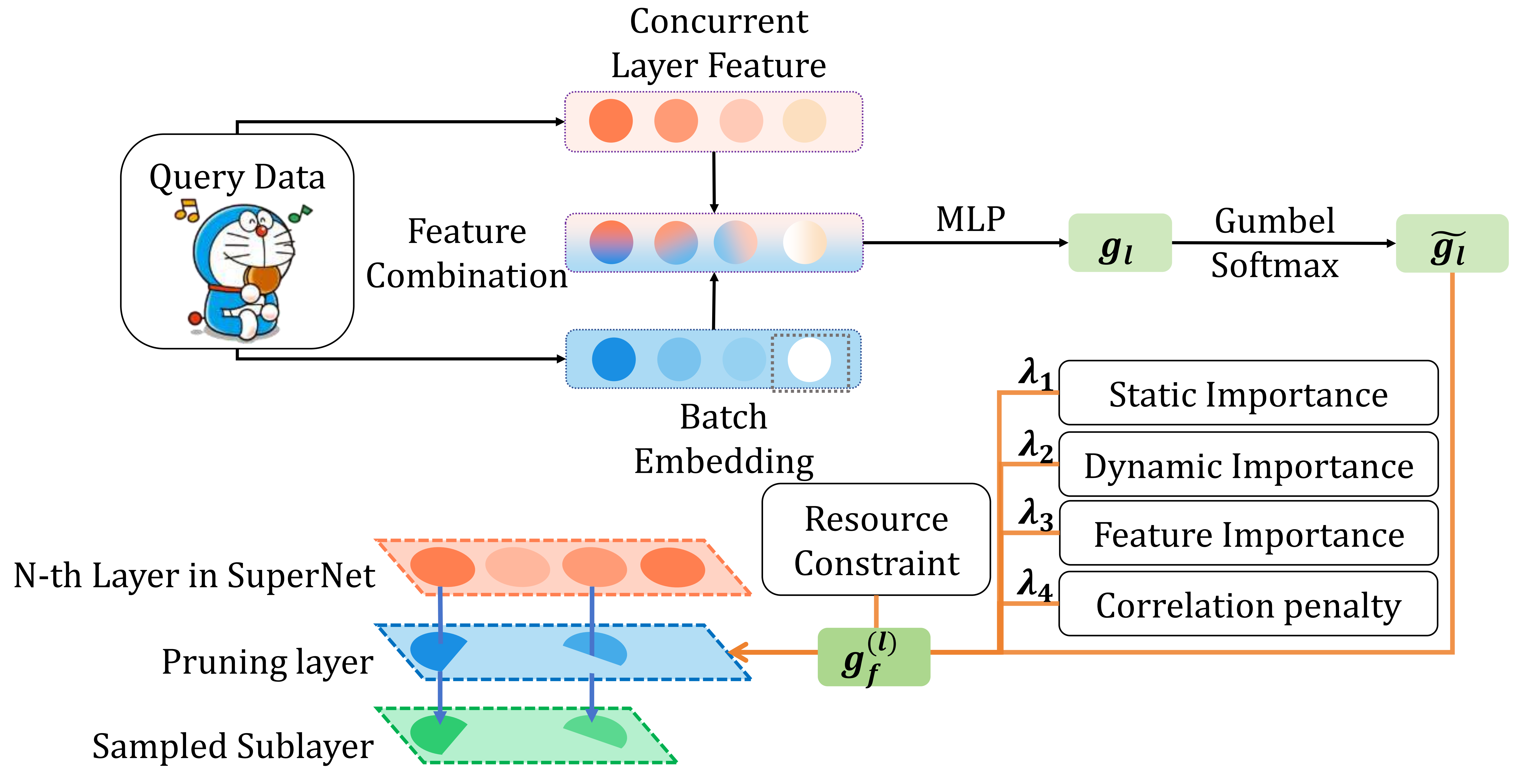}
\caption{Our layer-wise dynamic select gate mechanism. The framework combines concurrent layer features and batch embeddings through feature combination, applies gumbel-softmax for exploration, and generates final gates through sampling under resource constraints. The gate is also guided by four importance metrics evaluating static, dynamic, feature based, and correlation significance.}
\label{fig:gate1}
\end{figure}
Neural architecture search (NAS) has emerged as a promising approach for automating deep neural network design. However, current NAS methods face two key challenges in real-world deployments: (1) the static architectures lack flexibility to adapt to varying computational constraints, and (2) separate expensive searches are required for each deployment scenario, leading to prohibitive costs.
To address these limitations, we propose DANCE (Dynamic Architectures with Neural Continuous Evolution), which reformulates architecture search as a continuous evolution problem. Instead of finding a single fixed architecture, DANCE learns a distribution over architectural components that can be efficiently sampled and adapted to specific deployment requirements. This enables rapid architecture derivation while maintaining performance across diverse scenarios.
\begin{figure*}[t]
\centering
\includegraphics[width=\linewidth]{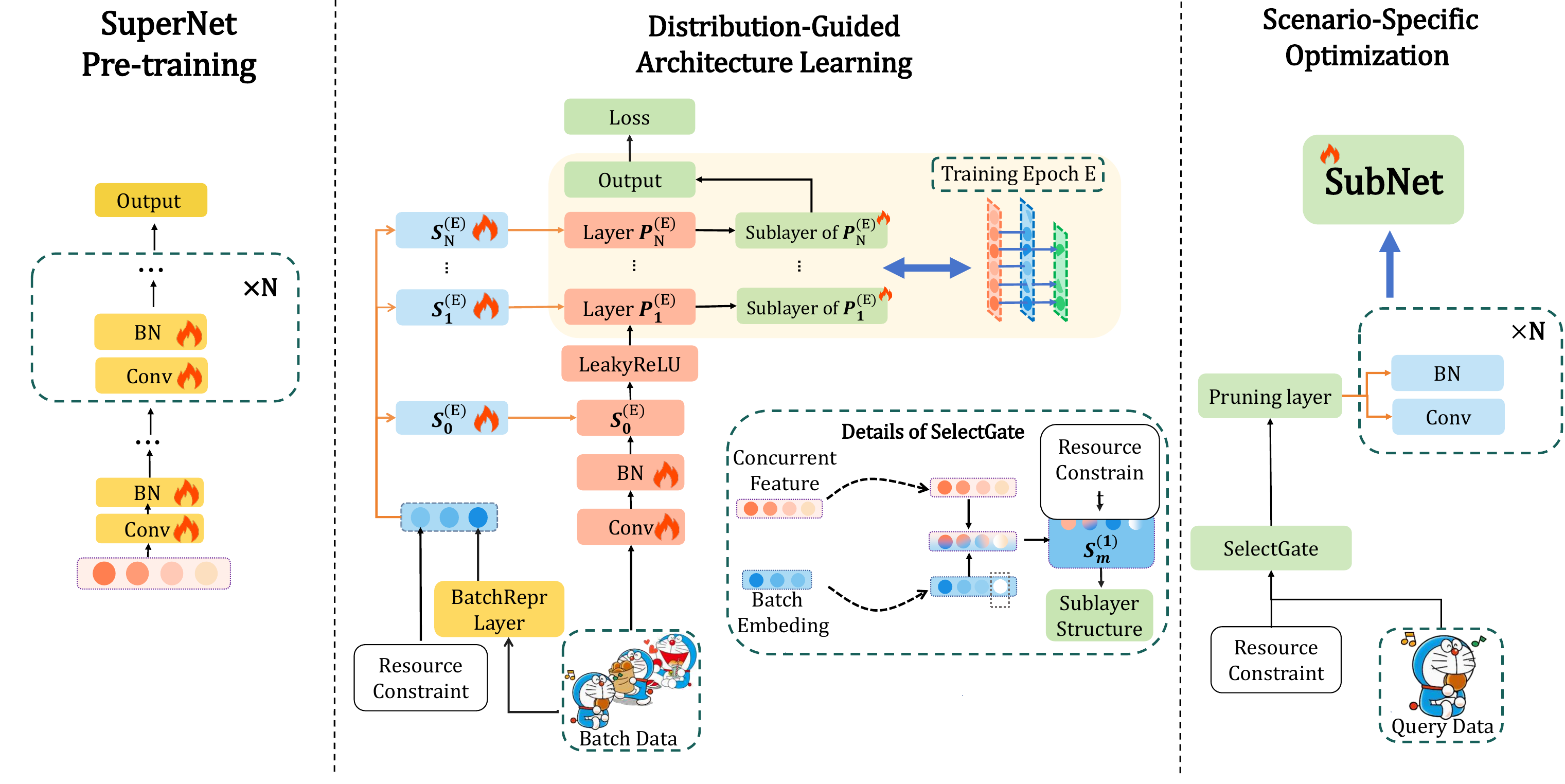}
\caption{Overview of DANCE's three-stage training process. Stage 1 pre-trains a SuperNet with N stacked layers as backbone. Stage 2 employs distribution-guided learning where SelectGate generates architecture masks based on batch features, layer features and constraints through dual-loop optimization. Stage 3 optimizes scenario-specific SubNets by inheriting SuperNet weights and fine-tuning under deployment requirements. This enables continuous adaptation while maintaining performance across diverse scenarios.}
\label{fig:gate}
\end{figure*}
\subsection{Problem Formulation}

Traditional NAS methods typically operate in a discrete search space, optimizing architectures under the fixed resource constraints:
\begin{equation}
\mathcal{A}^* = \arg\min_{\mathcal{A}} \mathcal{L}(\mathcal{A}) \text{ s.t. } \text{Cost}(\mathcal{A}) \leq C
\end{equation}
where $\mathcal{A}$ denotes a candidate architecture and $C$ specifies the computational budget constraint. This discrete optimization approach presents several challenges. First, it requires running separate searches for different deployment scenarios, leading to substantial computational overhead. Second, the discrete nature of the search space makes it difficult to adapt architectures smoothly as requirements change. Third, there is no guarantee of consistency between the selected components across different layers, resulting in suboptimal results.

\subsection{Unified Architecture Space Design}

Building upon the limitations of traditional NAS methods, DANCE introduces a unified architecture space design that accommodates diverse deployment requirements while significantly enhancing search efficiency. As shown in Figure~\ref{fig:gate1}, for each N-th SuperNet layer, we introduce an \textbf{data-aware dynamic select gate} to dynamically generate a pruning layer for component selection.

The key insight is learning a continuous distribution over architectural components that factorizes across layers while maintaining dependencies:
\begin{equation}
p(\mathcal{A}|\mathcal{D},C) = \prod_{l=1}^L p(g_l|\mathcal{F}_l,\mathcal{D},C)
\end{equation}

This formulation expresses the probability of selecting a specific architecture $\mathcal{A}$ as a product of layer-wise selection probabilities, where each layer's selection depends on the layer features $\mathcal{F}_l$, input data batch $\mathcal{D}$, and resource constraints $C_l$~(where $C_l$ represents the number of parameters to retain in layer $l$). Specifically, we implement this through a data-aware dynamic selection mechanism:

\begin{equation}
\begin{aligned}
g^{(l)}_{\text{f}} &= \text{SelectGate}(\mathcal{D}, \mathcal{F}_l,C_l), \quad l \in [1,L]
\end{aligned}
\end{equation}

For each n-th layer in the supernet, the \text{SelectGate} system combines concurrent layer feature and batch embedding through feature combination, which then passes through gumbel-softmax to generate select gate for layer selection. 

This process can be formulated as:

\begin{equation}
\begin{aligned}
\text{FeatureComb}_l &= \text{Combine}(\mathcal{D}, \mathcal{F}_l)\notag \\
g_l &= \text{MLP}(\text{FeatureComb}_l)
\end{aligned}
\end{equation}

where $\text{FeatureComb}_l$ is the combined feature and $g_l$ is the final select gate output. Here MLP is a two-layer neural network with ReLU activation in between, which learns to map the combined features to selection probabilities. Given the layer-wise resource budget $C_l$, the selection gate mechanism employs a combination of soft and stochastic selection to determine the final gate value.

\begin{equation}
\begin{aligned}
\epsilon &\sim \text{Gumbel}(0,1) \notag\\
\tilde{g}_l &= \text{Softmax}((g_l + \epsilon)/\tau) \notag\\
p_l &= \text{Softmax}(\tilde{g}_l \cdot \text{Score}(\mathcal{A}_l)) \notag\\
g^{(l)}_{\text{f}} &= \text{BernoulliSample}(p_l, C_l)
\end{aligned}
\end{equation}

Here, $\epsilon$ is sampled from a $Gumbel(0,1)$ distribution to introduce randomness in the softmax operation, $\tau$ is the temperature parameter controlling the sharpness of the softmax distribution, $\tilde{g}_l$ represents the soft selection probabilities after gumbel-softmax. $p_l$ normalizes the importance-weighted probabilities through softmax to ensure valid sampling probabilities. The $\text{BernoulliSample}$ function performs stochastic binary sampling where components are sampled according to their probabilities while ensuring exactly $C_l$ components are selected, maintaining the resource constraint while introducing controlled randomness in architecture exploration. $\text{Score}(\mathcal{A}_l)$ comprehensively evaluates component importance by considering static importance, dynamic importance, feature importance, and correlation penalty metrics. The resulting $g^{(l)}_{\text{f}}$ determines which components in the layer are retained or pruned, effectively generating a sublayer structure.
Detailed design and mathematical equations of score$(\mathcal{A}_l)$ are provided in \textbf{Appendix~Sec.A}.

\subsection{Multi-Stage Training Process}
To effectively learn and utilize the unified architecture space, DANCE employs a three-stage training process (as shown in Figure~\ref{fig:gate}) that progressively optimizes both the architectural distribution and model parameters at the same time.

\subsubsection{First Stage: SuperNet Pre-training} 
We first construct and pre-train a SuperNet containing N stacked layers as the backbone architecture. For each layer, we incorporate multiple parallel branches with different operators and channel dimensions to provide architectural flexibility. The SuperNet training process involves sampling different paths and optimizing them jointly to learn robust and transferable feature representations. To maintain stability during training, we employ progressive path dropping and knowledge distillation between different architectural configurations. Additionally, we introduce architecture-aware regularization to encourage diversity in the learned features across different components. This pre-training stage establishes a strong foundation that can effectively generalize across varying computational environments while ensuring consistent feature extraction capabilities. The pre-trained SuperNet serves as the starting point for subsequent distribution-guided architecture adaptation.
\subsubsection{Second Stage: Distribution-Guided Learning}    
For each layer $l$, in each batch iteration $t$, our SelectGate takes three key inputs: the batch-level distribution features $\mathcal{D}_t$, concurrent layer features $\mathcal{F}_l$, and layer-wise resource constraints $C_l$, generating binary masks $g_{\text{f}}^{(l,t)}$ to indicate components: $$g_{\text{f}}^{(l,t)} = \text{SelectGate}(\mathcal{D}_t, \mathcal{F}^l_t, C_l)$$

The training process employs a dual-loop optimization strategy that enables comprehensive architecture exploration while ensuring stable performance. In the outer loop across epochs $E$, the network progressively refines both its model weights and SelectGate parameters. 

In each batch iteration $t$, we sample $T$ different architectures using SelectGate to explore the architectural space while adhering to resource constraints. These sampled architectures are used to train the network weights, with SelectGate parameters being updated based on their performance feedback. This inner loop process enables thorough architecture exploration with efficient parameter optimization.

Specifically, for each batch iteration t, we first obtain T different architecture samples $\{\mathcal{A}^{(t)}_i\}_{i=1}^T$ following SelectGate's probability distribution under given resource constraints C. We then update both the network weights $\theta$ and SelectGate parameters $\phi$ using the training feedback from these sampled architectures. This dual-loop mechanism achieves both efficient architecture exploration and effective parameter optimization while satisfying deployment constraints.

\subsubsection{Third Stage: Scenario-Specific Optimization}
In the final stage, DANCE optimizes SubNets for specific deployment scenarios. Given the input query data $\mathcal{D}$ and resource constraints $C$, we obtain the optimal architecture $\mathcal{A}^*$ by inheriting weights from the SuperNet $\mathcal{W}$ and selecting features at each layer:

\begin{equation}
\begin{aligned}
g^{(l)} &= \text{SelectGate}(\mathcal{D}, \mathcal{F}_l,C_l ), \quad l \in [1,L] \\
\mathcal{A}^* &= \{g_l \odot \mathcal{W}_l\}_{l=1}^L
\end{aligned}
\end{equation}

where $g_l$ is the selection gate output at layer $l$ based on distribution $\mathcal{D}$ and features $\mathcal{F}_l$, and $\mathcal{W}_l$ represents the weights from layer $l$ of the SuperNet. The final architecture $\mathcal{A}^*$ is constructed by applying these pruning layers $g^{(l)}$ to the SuperNet weights.
We then fine-tune this architecture through a multi-objective optimization process task performance:
$
\min_{\theta} \mathcal{L}_{\text{total}}(\mathcal{A}^*) = \mathcal{L}_{\text{task}} 
$.
Through this carefully designed three-stage process, DANCE achieves continuous adaptation to varying computational constraints and data distributions, resource-efficient architecture search guided by distribution awareness, robust performance across diverse deployment scenarios, and end-to-end trainability without discrete sampling or complex optimization procedures.
\begin{algorithm}[t]
    \caption{Three-Stage Training Process}
    \textbf{Input}: Training data $\mathcal{D}$, Resource constraints $C$, Initial parameters $\theta$\\
    \textbf{Output}: Optimized SubNet $\mathcal{A}^*$
    \begin{algorithmic}[1]
    
        \STATE // Stage 1: SuperNet Pre-training
        \FOR{each training iteration}
            \STATE Sample batch data from $\mathcal{D}$
            \STATE Update SuperNet parameters $\theta$ with standard training
        \ENDFOR
        
        \STATE // Stage 2: Distribution-Guided Learning 
        \FOR{each batch iteration $t$}
            \STATE Extract features: $\mathcal{F}_l^t = \text{BatchRepr}_l(\mathcal{D}_t)$
            \FOR{each layer $l$}
                \STATE $g_{\text{f}}^{(l,t)} = \text{SelectGate}(\mathcal{D}_t, \mathcal{F}_l^t, C_l)$
            \ENDFOR
            \STATE Update parameters with sampled architectures
        \ENDFOR
        
        \STATE // Stage 3: Scenario-Specific Optimization
        \FOR{deployment scenario}
            \STATE Sample batch data from $\mathcal{D}$
            \STATE Generate architecture via SelectGate
            \STATE $\mathcal{A}^* = \{g^{(l)}_{\text{f}} \odot \mathcal{W}_l\}_{l=1}^L$
            \STATE Fine-tune $\mathcal{A}^*$ by minimizing $\mathcal{L}_{\text{total}}$
        \ENDFOR
        
        \RETURN $\mathcal{A}^*$
    \end{algorithmic}
\end{algorithm}

\subsection{Framework Analysis}

Given a network with $L$ layers and maximum width $W$ per layer, processing a batch of size $B$, DANCE achieves $\mathcal{O}(TBL(W + |C| + P))$ computational complexity, where $T$ is the number of training iterations, $|C|$ is the dimension of constraint embeddings, and $P$ denotes the number of trainable parameters in the network. Through distribution learning and continuous evolution, DANCE enables efficient architecture sampling at $\mathcal{O}(1)$ cost per scenario with the learned distribution $p(\mathcal{A}|\mathcal{D},C)$. While traditional NAS methods require separate searches for each deployment scenario leading to substantial computational overhead, DANCE learns a unified distribution that enables efficient architecture sampling for new scenarios. Compared to methods that directly evaluate architectures without training, DANCE's distribution learning captures both data characteristics and deployment constraints through $\mathcal{F}_l(\mathcal{D})$ for more accurate architecture generation. The discrete nature of traditional search spaces makes it difficult to adapt architectures smoothly as requirements change, but DANCE's continuous evolution through the selection gate $g_l$ enables smooth architecture adaptation via distribution sampling. Unlike methods requiring manual tuning with separate optimizations per scenario, DANCE automates this through the comprehensive scoring mechanism $\text{Score}(\mathcal{A})$ that maintains consistency between selected components across layers. The continuous evolution through the learned distribution $p(\mathcal{A}|\mathcal{D}, C)$ enables dynamic adaptation as constraints change, making DANCE particularly effective for deployment with varying computational requirements.

\section{Experiment}

To demonstrate the effectiveness of our proposed DANCE, we conduct extensive experiments on five image datasets.

 \begin{table*}[ht]
\centering
\resizebox{\textwidth}{!}{%
\begin{tabular}{l|l|cc|cc|cc|cc|cc}
\hline
\multirow{2}{*}{Network} & \multirow{2}{*}{Method} & \multicolumn{2}{c|}{Food-101} & \multicolumn{2}{c|}{Stanford Cars} & \multicolumn{2}{c|}{CUB-200} & \multicolumn{2}{c|}{CIFAR-100} & \multicolumn{2}{c}{CIFAR-10} \\
\cline{3-12}
& & Params & Acc & Params & Acc & Params & Acc & Params & Acc & Params & Acc \\
\hline
\multirow{8}{*}{ResNet18} & Original~(Pretrained) &11.69 & 85.01&11.69 &88.54 &11.69 &76.87&11.69 & 79.58&11.69&93.31 \\
& OFA-Width 0.85 &8.54 & 70.89&9.39&72.13 &8.08&61.10 &8.99&63.03&9.49&77.77\\
& OFA-Width 1.0 &11.70 & 72.60&11.27&79.23 &11.28 &66.60& 11.17&66.70& 11.17&78.23 \\
& FixMatch &8.89&75.90 & 9.01&79.02 & 9.02& 65.89& 8.93 &72.69 & 8.93 & 85.78\\
& RecNAS &8.80 & 77.50& 7.90& 72.39& 8.20&66.11 &8.30 & 66.50& 2.70&84.17 \\
& SPOS &7.30 &73.42 &7.20&70.48 &6.90 &60.54 &7.20& 60.76&6.40& \textbf{89.12}\\
& Ours (Stage 2) &2.50 & 75.41&2.80 & 76.02&3.00 & 64.68& 2.50& 70.92&2.10&78.48 \\
& Ours (Retrained) &\textbf{2.50} & \textbf{83.80}$^\ast$&\textbf{2.80} &\textbf{79.93}$^\ast$ &\textbf{3.00} &\textbf{66.90}$^\ast$&\textbf{2.50} & \textbf{78.37}$^\ast$& \textbf{2.10}&{87.19} \\
\hline
\multirow{8}{*}{VGG16} & Original~(Pretrained) & 37.78&78.62 &37.78 &81.89&37.78 &74.02 &37.78 &75.54&37.78 &93.87 \\
& OFA-Width 0.85 & 22.56&50.05 & \textbf{23.33}&64.43 & \textbf{22.04}&57.03 & \textbf{22.57}&59.36 & \textbf{21.34}& 65.91\\
& OFA-Width 1.0 & 37.88&56.54 & 37.70& 63.15&37.69 &59.93 &37.70& 60.07&37.25 &66.22 \\
& FixMatch & 30.22&58.89 & 30.22& 68.12&28.99 & 60.92&30.43 &68.30 & 30.22 & {81.78}\\
& RecNAS & 32.11& 50.64& 32.09&56.43 &31.40&55.29 &33.65 & 58.08& 33.65& \textbf{83.80}\\
& SPOS & 24.80&50.82 & 24.30&50.80 &24.80 &51.70&{25.50}& 52.60&22.50&72.19\\
& Ours (Stage 2) & 21.57& 60.02&26.12 & 72.80&25.96 & 69.76&26.06 & 73.72&25.70& 76.18\\
& Ours (Retrained) &\textbf{21.57}&\textbf{62.20}$^\ast$ &{26.12} &\textbf{ 76.62}$^\ast$&{25.96} &\textbf{72.83}$^\ast$ &{26.06} &\textbf{74.54}$^\ast$&{25.70} &{81.32} \\
\hline
\end{tabular}
}
\caption{Performance Comparison of Different Pruning Methods. The best results are in \textbf{bold} with marker $^\ast$ for statistical significance better than the second best except the original one ($p<0.05$).}
\label{tab:comparison}
\end{table*}
\subsection{Datasets and Backbones}
To systematically evaluate DANCE's adaptive capabilities and validate its technical innovations, we conduct comprehensive experiments across five datasets using two classic backbone architectures:
The evaluation framework employs CIFAR-10~\cite{krizhevsky2009learning} and CIFAR-100~\cite{krizhevsky2009learning} as foundational benchmarks to establish baseline architectural distributions. Three challenging fine-grained datasets—Stanford Cars~\cite{kramberger2020lsun}, CUB-200-2011~\cite{wah2011caltech}, and Food-101~\cite{bossard2014food}-are used to rigorously test the precision of dynamic selection mechanisms and architectural sampling under high-resolution visual discrimination requirements. These datasets test DANCE on several dimensions, including resolution scalability, architecture efficiency, and task complexity progression (basic categorization to fine-grained differentiation).  It demonstrates how DANCE can adapt to constantly changing task requirements while balancing resource constraints.
For more details about datasets and backbones, please refer to \textbf{Appendix~Sec.~B}.


\begin{table}[t]
    \centering
    \resizebox{0.48\textwidth}{!}{
    \begin{tabular}{c|ccccc}
    \toprule
    Ablation  & \multicolumn{5}{c}{Resource Constraint} \\
    Setting & 0.9 & 0.8 & 0.7 & 0.6 & 0.5 \\
    \midrule
    Default & 67.81 & 65.56 & 61.67 & 48.79 & 22.53 \\
    Only Static Importance  & 51.37 & 49.87 & 42.76 & 28.61 & 13.17 \\
    Only Dynamic Importance  & 39.90 & 36.76 & 28.14 & 14.08 & 10.58 \\
    Only Feature Importance & 45.21 & 41.96 & 32.42 & 16.13 & 14.10 \\
    Only Correlation Penalty  & 44.56 & 47.95 & 35.42 & 18.76 & 10.00 \\
    \bottomrule
    \end{tabular}
    }
       \caption{Ablation experiment on items of combined importance when pruning supernet.}
    \label{tab:ablation}
    \vspace{-3mm}
\end{table}

\subsection{Evaluation Metrics}
We establish a rigorous evaluation framework to assess the effectiveness of our proposed pruning method across multiple dimensions of performance and efficiency. The primary performance indicator is Top-1 accuracy measured across five datasets (CIFAR-10/100, Food-101, Stanford Cars, CUB-200), complemented by performance retention ratios relative to pretrained models. To quantify computational efficiency, we track both parameter reduction metrics and FLOPs measurements, analyzing pruning granularity through per-layer parameter changes and channel configuration modifications across network stages. We also performing detailed ablation studies to examine the contributions of static, dynamic, and feature importance components. The framework incorporates cross-platform evaluation across different model scales (ResNet-18~\cite{he2016deep}, VGG-16~\cite{simonyan2014very}) and progressive assessment under varying resource constraints (target sparsity from 0.5 to 0.9), enabling comprehensive analysis of the accuracy-efficiency trade-off. The stability and adaptability of our method are further validated through fine-grained resource constraint variations (0.1 to 0.5 with 0.02 step size), providing insights into performance consistency across diverse deployment scenarios. This comprehensive evaluation methodology enables us to rigorously validate our method's capability in achieving optimal balance between model performance and resource efficiency while maintaining robust adaptability.

\subsection{Baselines}
To demonstrate the effectiveness and efficiency of our approach, we compare our approach with different state-of-the-art NAS methods. OFA~\cite{cai2019once} uses progressive shrinking to train a supernet that supports diverse architectural configurations, enabling flexible deployment under different resource constraints. ProxylessNAS~\cite{cai2018proxylessnas} directly trains specialized architectures on target hardware while addressing resource constraints through gradient-based optimization. RecNAS~\cite{peng2022recnas} employs recursive channel pruning to find efficient architectures, achieving a better trade-off between accuracy and model size. SPOS~\cite{guo2020single} adopts a single-path one-shot architecture to train a supernet, allowing efficient architecture sampling and evaluation. FixMatch~\cite{sohn2020fixmatch} leverages a semi-supervised learning~\cite{zhu2005semi} strategy to improve model performance with limited labeled data through consistency regularization. The key difference in our approach is the integration of dynamic channel selection and correlation-aware pruning, which enables more effective identification of redundant parameters while maintaining model accuracy across different sparsity levels.
We also implement a Score-based Pruning baseline that directly applies our architecture evaluation metric Score($\mathcal{A}$) (defined in ~\textbf{Appendix Sec.A} and \textbf{Sec 2.2}) for the direct parameter.  

\subsection{Implementation Details}
When implementing the training process, we adopt a two-stage approach with carefully selected hyperparameters. Stage 1(Pre-training)runs for epochs with frozen SelectGate modules, while Stage 2 (Distribution-Guided Architecture Learning) continues for epochs with all components activated. The learning rates are selected from [0.0001, 0.0005, 0.001] for different components. We employ the AdamW optimizer with OneCycleLR scheduler using 30\% warm-up period and early stopping patience of 15.
All components are implemented using PyTorch and monitored through comprehensive metrics, including accuracy, loss components, and gate statistics. The backbone network uses ResNet-18 and VGG-16 architecture with customized SelectGate modules integrated at different layers. Training is performed on standard GPU hardware with automatic mixed precision (AMP) enabled for efficiency. For more details about datasets and backbones, please refer to \textbf{Appendix~Sec.~C}.
\subsection{Overall Performance Comparison}
As shown in Table~\ref{tab:comparison}, we observe three remarkable phenomena across five datasets. First, our Stage 2 results achieve strong performance through direct sampling without any retraining - notably outperforming several baseline methods that require full fine-tuning. For instance, on CIFAR-100, our Stage 2 sampling achieves $70.92\%$ accuracy, surpassing RecNAS ($66.50\%$) and SPOS ($60.76\%$). Second, with minimal parameters ($2.1$M-$3.0$M for ResNet-18), our method maintains competitive accuracy across diverse tasks. Third, while retraining further boosts performance, the strong Stage 2 results already demonstrate the effectiveness of our approach.

These phenomena stem from our distribution-guided architecture learning strategy. Unlike traditional methods that require separate, expensive fine-tuning for each deployment scenario, our \text{SelectGate} mechanism learns to directly generate effective architectures by comprehensively evaluating both the static and dynamic importance of components. The data-aware nature of our approach enables the \text{SelectGate} to capture task-specific requirements and generate suitable architectures through simple sampling from the learned distribution $p(\mathcal{A}|\mathcal{D},C)$. 
The exceptional Stage 2 performance validates our core insight of reformulating architecture search as a continuous evolution problem. Our approach enables rapid architecture derivation through direct sampling while maintaining strong performance by learning a distribution over architectural components rather than searching for fixed architectures. This represents a significant advantage over existing methods that rely heavily on costly fine-tuning procedures.

\begin{figure}[t]
    \centering
    \includegraphics[width=0.95\linewidth]{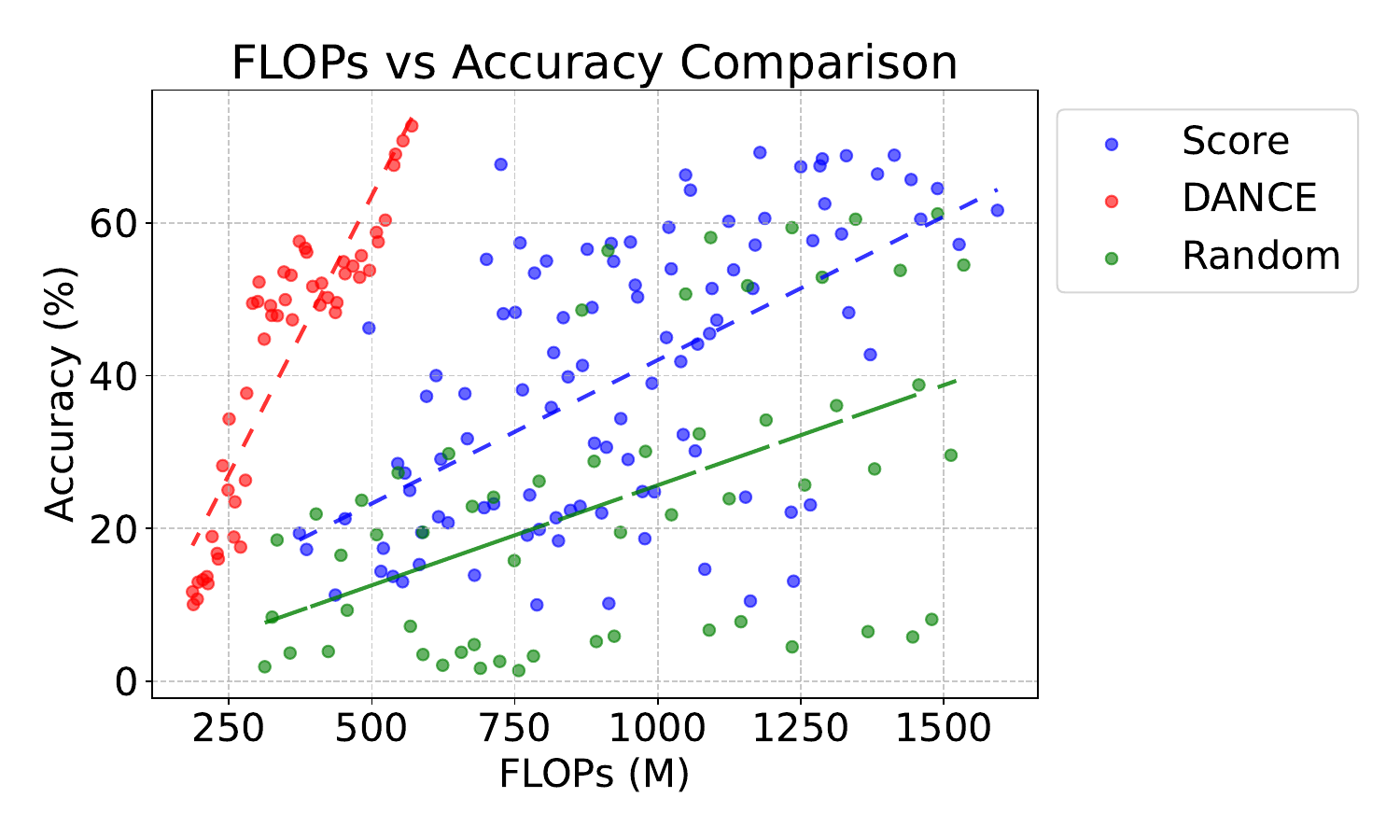}
    \caption{Distribution of FLOPs vs. Accuracy across different pruning strategies. DANCE achieves higher accuracy with lower FLOPs compared to score-based pruning, demonstrating a better efficiency-accuracy trade-off. Random architecture selection on SuperNEt, which randomly samples network architectures, performs significantly worse. The dashed lines show the general trends.}
    \label{fig:flops_accuracy}
    \vspace{-3mm}
\end{figure}

\subsection{Effectiveness of Distribution-Guided Learning vs Simple Score-Based Sampling}
As shown in Figure~\ref{fig:flops_accuracy}, we observe several key advantages of DANCE over score-based pruning, despite both methods utilizing the same importance scoring mechanism. First, DANCE achieves substantially better efficiency-accuracy trade-off, with a much steeper slope in the accuracy-FLOPs curve. While score-based pruning requires 1200-1500M FLOPs to reach 60\% accuracy, DANCE achieves similar or better accuracy with only 400-500M FLOPs, demonstrating superior efficiency. Second, score-based pruning shows scattered and inconsistent performance (blue dots widely spread), while DANCE demonstrates stable and predictable behavior with a clear trend (red dots closely following the trend line). Random architecture selection on SuperNet performs significantly worse, validating the importance of structured pruning strategies.
These phenomena validate that merely using importance scores for selection is insufficient. Although both methods leverage the same scoring mechanism, our \text{SelectGate}'s distribution-guided learning strategy enables more effective architecture generation by learning to combine and utilize the importance scores through the continuous distribution $p(\mathcal{A}|\mathcal{D},C)$. The substantial performance gap demonstrates that the key advantage comes from our distribution-guided learning framework rather than the scoring mechanism itself.

\begin{figure}[t]
    \centering
    \includegraphics[width=0.95\linewidth]{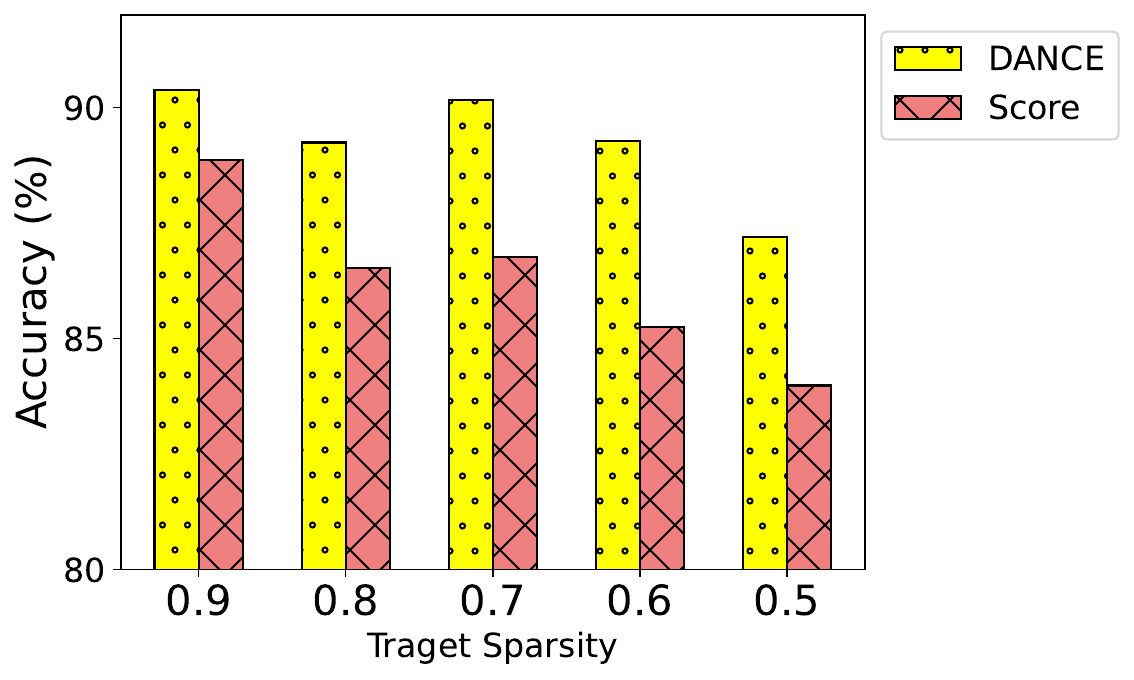}
    \caption{Performance comparison between DANCE and Score-based Purning under different sub-network constraints. Our method consistently demonstrates better performance across different scales, validating the effectiveness of the distribution-guided search.}
    \label{fig:subnetwork_comparison}
\end{figure}

To further validate this finding, we conduct experiments on sub-networks with different constraints. As shown in Figure~\ref{fig:subnetwork_comparison}, DANCE consistently outperforms score-based pruning across different scales. This consistent superior performance validates the effectiveness of our distribution-guided learning framework, showing its ability to maintain high accuracy while satisfying various computational constraints.

\subsection{Ablation Study}
Our sampling strategy incorporates four essential and complementary dimensions of importance metrics. Static importance measures the inherent significance of nodes by reflecting their fundamental structural contributions, while dynamic importance captures runtime activation patterns that reveal operational importance during inference. Feature importance evaluates the discriminative power of different features to preserve critical information, and the correlation penalty balances these metrics by reducing redundancy for efficient sampling. As shown in Table~\ref{tab:ablation}, using only a single importance metric leads to substantial performance drops across all resource constraints, validating that these dimensions must work in synergy to achieve optimal sampling effectiveness.
Hyperparameter analysis can be found at ~\textbf{Appendix~Sec .~D}.
\section{Related Work}

\noindent\textbf{Neural Architecture Search.}
Neural Architecture Search (NAS) aims to automate the discovery of optimal neural network architectures~\cite{elsken2019neural}. Early NAS methods relied on reinforcement learning or evolutionary algorithms to explore discrete architecture spaces, but suffered from prohibitive computational costs due to training each candidate architecture from scratch~\cite{zoph2016neural}. One-shot NAS methods~\cite{pham2018efficient} improved efficiency by training a supernet containing all possible architectures and sharing weights between candidates. While this reduced search costs, these methods still struggle with effectively adapting architectures across different deployment scenarios due to their discrete nature, as they primarily focus on block-level architecture components rather than feature dimensions.

Recent works have explored more flexible approaches. Progressive NAS~\cite{liu2018progressive} gradually increases architecture complexity during search but lacks mechanisms for handling varying computational constraints. Differentiable NAS methods~\cite{wang2021rethinking} enable end-to-end architecture optimization but face challenges in maintaining performance consistency across diverse deployment contexts. Although these advances have improved search efficiency, they do not fully address the need for dynamic architecture adaptation under varying computational requirements.

DANCE demonstrates consistent accuracy gains and reduced search costs across diverse datasets, proving its effectiveness for practical applications with varying computational demands and resource limitations.

\noindent\textbf{Efficient Architecture Design.}
The growing demand for deploying deep neural networks across different computational environments has sparked interest in efficient architecture design~\cite{cai2018efficient}. Traditional network pruning approaches focus on removing redundant parameters or channels based on importance metrics~\cite{han2015learning}, while NAS methods typically search over predefined block-level components. Our work bridges this gap by reformulating both pruning and NAS from a unified feature dimension perspective, enabling more flexible architecture adaptation.

More recent approaches have focused on automated, efficient architecture design. Resource-aware NAS methods~\cite{yang2021efficient} explicitly incorporate hardware constraints during search but struggle with smooth architecture adaptation as requirements change. While these methods have shown promise for specific deployment contexts, they lack mechanisms for effectively modeling the complex trade-offs between multiple competing objectives across diverse scenarios. Additionally, their block-based search spaces limit the granularity of architecture optimization compared to our feature-dimension based approach.

Our DANCE framework addresses these limitations by unifying NAS and pruning through continuous evolution over feature dimensions rather than discrete blocks. This novel perspective enables more efficient and flexible architecture adaptation while maintaining performance consistency across diverse deployment scenarios. 

\section{Conclusion}

This paper presents DANCE, a novel framework that reformulates neural architecture search through continuous evolution over feature dimensions rather than traditional block-level components. Our approach uniquely bridges the gap between NAS and pruning, providing a unified perspective for flexible architecture adaptation across diverse deployment scenarios. Through extensive experiments, we demonstrate that DANCE successfully addresses several key limitations of existing methods by enabling smooth architecture adaptation, effectively handling diverse computational constraints, and efficiently balancing multiple competing objectives during the search process.
Our results show that DANCE achieves superior accuracy compared to state-of-the-art methods while significantly reducing search costs across various real-world deployment scenarios. These findings validate the effectiveness of viewing architecture search through the lens of feature-dimension optimization and continuous evolution, suggesting a promising direction for future research in automated neural architecture design.

\section*{Acknowledgements}
This research was partially supported by Research Impact Fund (No.R1015-23), Collaborative Research Fund (No.C1043-24GF), Huawei (Huawei Innovation Research Program, Huawei Fellowship), Tencent (CCF-Tencent Open Fund, Tencent Rhino-Bird Focused Research Program), Alibaba (CCF-Alimama Tech Kangaroo Fund No. 2024002), Ant Group (CCF-Ant Research Fund), and Kuaishou.

\section*{Contribution Statement}
Maolin Wang and Tianshuo Wei conceptualized the research idea, designed the methodology, and conducted the primary experiments. Maolin Wang led the software implementation, manuscript writing and data analysis. Tianshuo Wei contributed to software implementation and validation. Ruocheng Guo (corresponding author)  provided critical insights on the experimental design and data interpretation. Wanyu Wang (corresponding author) supervised the research direction and provided overall guidance. Shanshan Ye (corresponding author) contributed to the methodology verification and oversaw the experimental validation. Sheng Zhang assisted with data collection and analysis, and contributed to manuscript preparation. Lixin Zou assisted with data analysis and visualization. Xuetao Wei provided valuable feedback on the manuscript and contributed to the discussion section. Xiangyu Zhao supervised the project, secured funding, and provided critical revision of the manuscript. All authors reviewed and approved the final version of the manuscript.
\bibliographystyle{named}
\bibliography{main}

\section*{\Large{Appendix of DANCE}}
\section*{A. Details of Component Scoring Mechanism}

The Score($\mathcal{A}_l$) function establishes a theoretically grounded framework for evaluating architectural components. Our key insight is that component importance should be assessed from multiple complementary perspectives, each capturing distinct aspects of architectural significance. We propose a unified scoring mechanism that combines four carefully designed metrics, each with specific theoretical motivations and architectural implications.

The first metric addresses the inherent importance of components through static importance measurement:
\begin{equation*}
\begin{aligned}
\mathcal{I}_{\text{static}} &= \sigma(\theta_c), \quad \text{where}~\theta_c \sim \mathcal{N}(0, 0.01)
\end{aligned}
\end{equation*}

This formulation is motivated by the hypothesis that each architectural component possesses an intrinsic importance independent of input data. The sigmoid activation $\sigma(\cdot)$ is specifically chosen to normalize importance scores while maintaining differentiability, enabling end-to-end optimization. The normal initialization ($\mathcal{N}(0, 0.01)$) reflects our prior assumption that components should initially have similar importance, allowing the network to learn meaningful importance distributions through training.

To capture the dynamic nature of component importance in response to varying input patterns, we introduce a Gumbel-Softmax~\cite{jang2016categorical} based mechanism:
\begin{equation*}
\begin{aligned}
& g_t = \sigma(\text{fusion}(f_t) + \theta_c + \lambda_1\epsilon), \quad \epsilon \sim \text{Gumbel}(0,1) \\&
\mathcal{I}_{\text{dynamic}} = (1-\alpha)\mathcal{I}_{t-1} + \alpha\mathbb{E}_B[g_t]
\end{aligned}
\end{equation*}

This formulation is inspired by two key insights: First, component importance should adapt to input distributions, which is achieved through the fusion function combining current feature information with learnable parameters. Second, the architecture selection process should maintain some level of exploration during training, implemented through the Gumbel-Softmax reparameterization. The EMA update mechanism ($\mathcal{I}_{\text{dynamic}}$) reflects our assumption that importance should evolve smoothly over time while remaining responsive to distribution shifts.

For evaluating learned representations, we propose a feature-based importance metric:
\begin{equation*}
\begin{aligned}
f_t &= \text{FeatureNet}(x) \\
s_t &= \text{softmax}(\mathbb{E}_B[f_t]) \\
\mathcal{I}_{\text{feature}} &= (1-\alpha)\mathcal{I}_{t-1} + \alpha s_t
\end{aligned}
\end{equation*}

This design is based on the principle that effective architectural components should learn discriminative feature representations. The softmax normalization implements a competitive mechanism among components, reflecting our hypothesis that relative feature importance is more meaningful than absolute measures. The EMA update balances stability and adaptivity in feature importance estimation.

To promote architectural diversity, we introduce a correlation-based penalty:
\begin{equation*}
\begin{aligned}
R &= \text{corrcoef}(A^T) \\
\mathcal{I}_{\text{corr}} &= 1 - \lambda_2\mathbb{E}[|R|]
\end{aligned}
\end{equation*}

This term is motivated by the fundamental principle that optimal architectures should minimize redundancy among components. The correlation coefficient captures linear dependencies between activation patterns, while the penalty term encourages the selection of complementary components. The absolute value operation ensures both positive and negative correlations are penalized, reflecting our assumption that truly complementary components should exhibit independent activation patterns.

These four theoretically motivated metrics are combined through a weighted summation:
\begin{equation*}
\begin{aligned}
\text{Score}(\mathcal{A}_l) = \lambda_1\mathcal{I}_{\text{static}} + \lambda_2\mathcal{I}_{\text{dynamic}} + \lambda_3\mathcal{I}_{\text{feature}} + \lambda_4\mathcal{I}_{\text{corr}}
\end{aligned}
\end{equation*}

This unified scoring mechanism reflects our core hypothesis that effective architecture evaluation requires balancing multiple complementary aspects of component importance. The weighting coefficients $\lambda_1$ through $\lambda_4$ provide flexibility in emphasizing different importance aspects based on specific architectural requirements. The static importance ($\mathcal{I}_{\text{static}}$) maintains architectural stability, while dynamic importance ($\mathcal{I}_{\text{dynamic}}$) enables adaptation to input distributions. Feature importance ($\mathcal{I}_{\text{feature}}$) ensures the selection of components that learn meaningful representations, and correlation penalty ($\mathcal{I}_{\text{corr}}$) promotes architectural diversity. This theoretically grounded approach enables both efficient architecture search and optimal component selection while maintaining interpretability and architectural diversity.

\begin{figure*}[t]
    \centering
    \begin{subfigure}{0.24\linewidth}
        \centering
        \includegraphics[width=\linewidth]{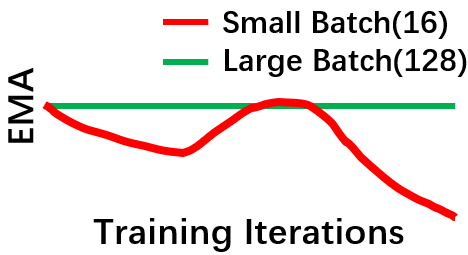}
        \caption{Batch size analysis}
    \end{subfigure}
    \hfill
    \begin{subfigure}{0.24\linewidth}
        \centering
        \includegraphics[width=\linewidth]{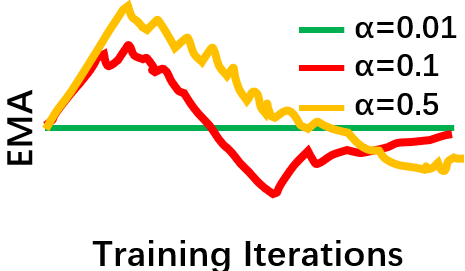}
        \caption{$\alpha$ parameter}
    \end{subfigure}
    \hfill
    \begin{subfigure}{0.24\linewidth}
        \centering
        \includegraphics[width=\linewidth]{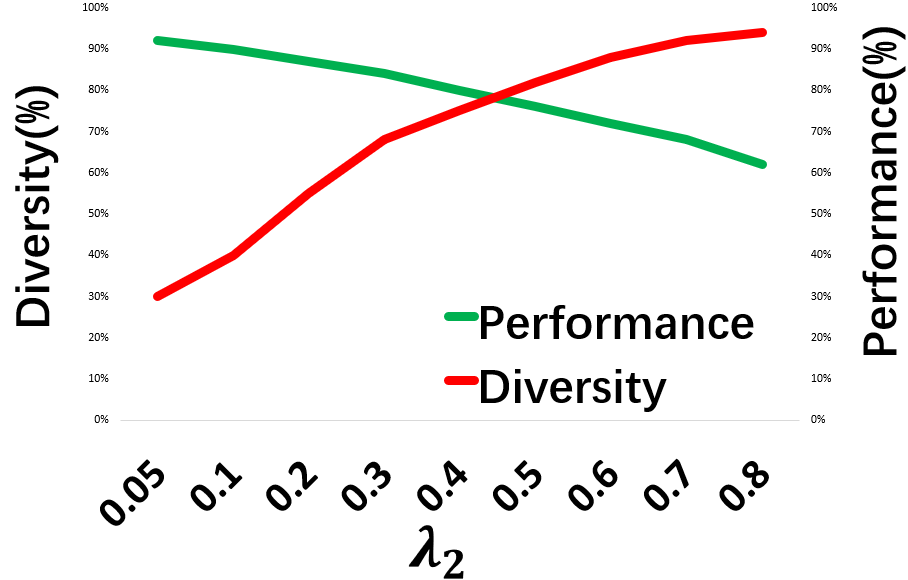}
        \caption{$\lambda_2$ parameter}
    \end{subfigure}
    \hfill
    \begin{subfigure}{0.24\linewidth}
        \centering
        \includegraphics[width=\linewidth]{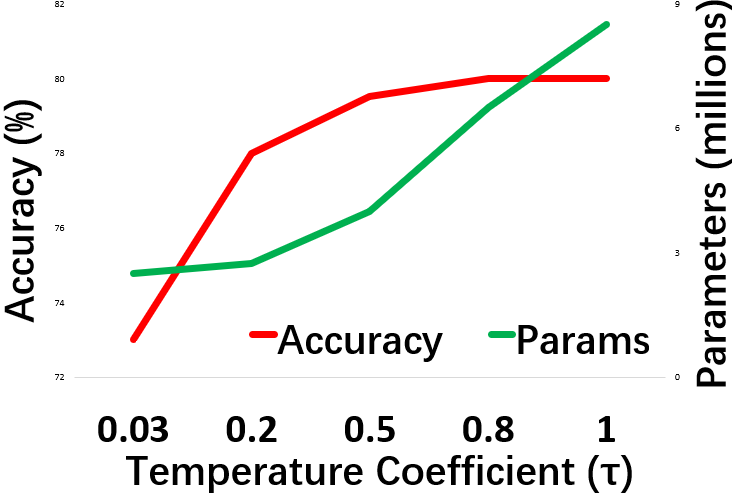}
        \caption{Temperature $\tau$}
    \end{subfigure}
    \caption{Parameter sensitivity analysis}
    \label{fig:parameters}
\end{figure*}
\section*{B. Datasets and Backbones}
As mentioned, in order to verify the feasibility and effectiveness of DANCE. We used 5 datasets (Table~\ref{tab:datasets}) and 2 backbones (ResNet-18~\cite{he2016deep} and VGG-16~\cite{simonyan2014very}). 
\begin{table}
\centering
\caption{Experimental Datasets}
\label{tab:datasets}
\begin{tabular}{l|cccc}
\hline
Dataset & Train & Test & Class & Resolution \\
\hline
Cifar-10 & 50000 & 10000 & 10 & 32 \\
Cifar-100 & 50000 & 10000 & 100 & 32 \\
Cub-200 & 5994 & 5794 & 200 & 224 \\
Stanford-Cars & 8144 & 8041 & 196 & 224 \\
Food-101 & 75750 & 25250 & 101 & 224 \\
\hline
\end{tabular}
\end{table}

CIFAR-10/100~\cite{krizhevsky2009learning}: A basic computer vision dataset with small (32×32) images. CIFAR-10 has 10 classes with 6,000 images each, while CIFAR-100 has 100 classes with 600 images each. Commonly used as an entry-level benchmark.
CUB-200~\cite{wah2011caltech}: A specialized dataset of 11,788 high-resolution (224×224) bird images covering 200 species. Challenges AI to distinguish subtle differences between similar-looking birds.
Stanford Cars~\cite{kramberger2020lsun}: Contains 16,185 car images (224×224) of 196 different models. Tests AI's ability to identify minor variations between car models and years.
Food-101~\cite{bossard2014food}: Features 101,000 real-world food photos across 101 categories. Challenges include varying image quality, lighting, and angles, plus distinguishing similar food types.

From the Table~\ref{tab:datasets} for Cifar-10/100 ratio between the Train and Test is 5:1. And for Cub-200 and Stanford-Cars satasets the ratio is almost 1:1. Ratio is 3:1 for Food-101. When training we will take 10\% of train set as valid set.

\section*{C. More Details about Implementation Details}

When implementing the training process, we adopt a three-stage approach with carefully selected hyperparameters:

Stage 1 (Representation Learning, 100 epochs) focuses on pre-training the SuperNet backbone with frozen SelectGate modules. During this stage:

Standard supervised training with cross-entropy loss
Basic data augmentation including random crop and horizontal flip
AdamW optimizer (lr=0.001) with OneCycleLR scheduler (30\% warm-up)
Early stopping with patience=15
Stage 2 (Joint Optimization, 100 epochs) activates SelectGate modules for distribution-guided learning:

Learning rates selected from [0.0001, 0.0005, 0.001] for different components. Three separate AdamW optimizers for gates, batch representation, and backbone. Loss function combines classification, sparsity (with dynamic weights), diversity, correlation, and stability terms. Layer-wise sparsity constraint $C_l$ selected from [0.3, 0.4, 0.5]
Gumbel-Softmax~\cite{} temperature chosen from [0.1, 0.5, 1.0]
Stage 3 (Deployment Optimization) generates scenario-specific SubNets by:

Inheriting trained SuperNet weights
Fine-tuning under specific deployment constraints
Adapting to target hardware requirements
The entire framework is implemented using PyTorch with automatic mixed precision (AMP) enabled. SelectGate modules utilize feature extraction networks (kernel size=3) and batch-level representations. Training progress is monitored through comprehensive metrics including accuracy, loss components, and gate statistics. For a typical 64×64 linear layer with $C_l$=0.5, the resulting 32×32 structure achieves over 50\% parameter reduction. Training is performed on standard GPU hardware.

\section*{D. Parameter Analysis}

 From Figure~\ref{fig:parameters}, larger batch sizes provide more stable updates and consistent architecture decisions while adjusting the $\mathbf{\alpha}$ parameter balances between responsiveness (high $\mathbf{\alpha}$) and stability (low $\mathbf{\alpha}$). We also dynamically tune $\mathbf{\lambda_2}$ from low values that promote exploration to higher values that optimize architecture, effectively balancing diversity and performance under different constraints. Temperature $\tau$ presents a trade-off: higher values improve accuracy but lead to larger model sizes, with diminishing returns in performance.
\end{document}